# IEDM, an Ontology for Irradiation Experiment Data Management

Blerina Gkotse[1,2], Pierre Jouvelot[1] and Federico Ravotti[2]

[1] MINES ParisTech, PSL University, Paris, France
[2] Experimental Physics Department, CERN, Geneva 23, CH-1211, Switzerland
`Blerina.Gkotse@cern.ch`

**Abstract.** Irradiation experiments (IE) are an essential step in the development of High-Energy Physics (HEP) particle accelerators and detectors. They assess the radiation hardness of materials used in HEP experimental devices by simulating, in a short time, the common long-term degradation effects due to their bombardment by high-energy particles. IEs are also used in other scientific and industrial fields such as medicine (e.g., for cancer treatment, medical imaging, etc.), space/avionics (e.g., for radiation testing of payload equipment) as well as in industry (e.g., for food sterilization). Usually carried out with ionizing radiation, these complex processes require highly specialized infrastructures: the irradiation facilities. Currently, hundreds of such facilities exist worldwide.

To help develop best practices and promote a computer-assisted handling and management of IEs, we introduce IEDM, a new OWL-based Irradiation Experiment Data Management ontology. This paper provides an overview of the classes and properties of IEDM. Since one of the key design choices for IEDM was to maximize the reuse of existing foundational ontologies such as the Ontology of Scientific Experiments (EXPO), the Ontology of Units of Measure (OM) and the Friend-of-a-Friend Ontology (FOAF), we discuss the methodological issues of the integration of IEDM with these imported ontologies. We illustrate the use of IEDM via an actual IE recently performed at IRRAD, the CERN proton irradiation facility. Finally, we discuss other motivations for this work, including the use of IEDM for the generation of user interfaces for IE management, and their impact on our methodology.

## 1 Introduction

In an irradiation experiment (IE), a piece of material, or even an entire device, is purposefully submitted to the waves or particles emitted by a radiative process; the time of exposure, the energy and the type of radiation field (gamma rays, X-rays, protons, ions, etc.) are parameters of the experiment. IEs are key experimental-physics procedures required, for instance, when designing, developing or building High-Energy

Physics (HEP) accelerators and experiments. In the HEP field, the purpose of performing an IE is typically the qualification of materials, detectors and electronic components in a radiation environment equivalent to the one these devices will encounter in the actual HEP experiment, thus simulating in a short time long-term radiation-induced degradation effects [1]. Moreover, worldwide, IEs also find applications in other scientific and technical fields. For example, in the field of space and avionics, the materials composing aircraft or spaceships also suffer from radiation damage during their flights. Therefore, engineers need to test space components and materials during their development projects [2]. Other examples can be found in industry, where irradiation experiments are used for various purposes such as for food sterilization [3] or seed treatment [4]. IEs are also nowadays performed on patients as part of radiotherapy treatments [5] or for medical imaging in many hospital environments [6]. Even though all these IEs are performed for different purposes and in various locations (e.g., hospitals, factories, scientific institutes), these infrastructures can all be assimilated, at the abstract level, to what we call "irradiation facilities".

An earlier thorough survey about the existent irradiation facilities of interest for HEP applications and their practices shows that at least hundreds of these are operational around the world [7]. Considering all the other IE-related fields of applications mentioned above, we estimate that these are only a small subset of what is currently existing worldwide. Although these facilities use and produce knowledge of high scientific value, our survey shows also that most of the teams in charge of the operation of IEs follow informal procedures for their overall data handling; they often store their data in ad-hoc spreadsheets in local computers or online, and even sometimes only on paper. This practice makes irradiation facilities more prone to the risk of data errors, corruption or even loss, hence the apparent need for the introduction of a standardized approach in the data management of IEs.

Ontologies are known to facilitate knowledge sharing, reusability, and formalization of specific domains [8]. Therefore, in this work, we decided to formalize the general knowledge linked to the management of data associated with IEs by introducing a new ontology, called Irradiation Experiment Data Management (IEDM) ontology. IEDM has been built by investigating and analyzing the elements and practices commonly used in irradiation facilities around the world, building thus upon the knowledge of domain experts to provide a first step towards the structured management of IE data.

Developing an ontology is a long and incremental process that calls for significant testing along the way. In order to provide the first evidence of the validity of our model, we illustrate in this paper how an actual IE can be described using IEDM. We chose FCC-Radmon [9], an experiment that has been performed at the CERN (European Laboratory for Particle Physics [10]) IRRAD proton irradiation facility [11] which is linked to the current development of a future particle monitoring device for the proposed Future Circular Collider (FCC) [12] under consideration at CERN.

The development of IEDM is part of a larger project undertaken at CERN and linked to the management of IE data [13]. In addition to the previous motivation related to domain formalization, we intend to use IEDM as a model from which a whole suite of web-based applications dedicated to IE data management can be automatically generated. Even further down the line, IEDM could be the foundational stone of an ontology for all irradiation facilities. These long-term goals, detailed in this paper, have had a significant impact on the way we structured IEDM, thus requiring specific methodological stances, which are described in the sequel. In particular, the development of IEDM relies heavily on existing open-source standards such as Web Ontology Language (OWL) [14] or physics-related ontologies (see Section 2).

The contributions of this paper are:
1. IEDM, a new OWL-based ontology dedicated to the management of data related to IEs;
2. an experimental validation of IEDM with the encoding of FCC-Radmon, an actual irradiation experiment performed in 2018 at the CERN IRRAD proton irradiation facility;
3. key methodological principles for ontology development linked to our planned use of ontology-based representation data for the automatic code generation of IE Web applications.

The present paper is structured as follows. In Section 2, we investigate the current state of the art in physics-related ontology development. Section 3 describes the methodology and techniques we have followed for the development of the IEDM ontology. In Section 4, we provide an overview of the main classes and relations present in IEDM. In Section 5, we detail our test encoding in IEDM of an actual experiment, FCC-Radmon. In Section 6, we discuss future work before, finally, presenting our conclusions in Section 7.

## 2 Related Work

In the literature, several ontologies formalize and axiomatize the knowledge of physics. One example of these is the Web Physics Ontology [15], which describes physics equations and relationships among physical quantities. However, this ontology is limited to the domain of electromagnetism and mechanics, and there is no reference to particle physics. Another approach is the ontology design pattern proposed for the particle physics analysis [16]. Although this work includes concepts that are typical in HEP experiments, it focuses only on the analysis of HEP data and not the representation of experiments.

To the best of our knowledge, an ontology dedicated to the formalization of the principles of irradiation experiments does not exist yet; the goal of our paper is to describe IEDM, an ontology that we specifically designed to address such a concern. To promote best practices, we reuse as much as possible the earlier ontology developments that we believe can partially describe irradiation experiments. The ones that we integrated in our ontology are described in the following subsections.

### 2.1 EXPO

The Ontology of Scientific Experiments (EXPO) is a general ontology for the formalization of scientific experiments. It introduces concepts specifically linked to the notions of experimental design, scientific methods, and other core principles of experiments [17]. Inheriting some of its structure from the Suggested Upper Merged Ontology (SUMO) [18], EXPO similarly classifies all its concepts into an abstract level and a physical one, providing in this way an elegant and quite flexible design. Taking both advantage of the SUMO-inspired abstract/physical structure and of the fundamental science-related entities of EXPO, IEDM derives many of its key classes from these two upper-level ontologies.

### 2.2 OM

However, since EXPO does not elaborate on physical quantities and units, restricting its key constructs to the logical structure of scientific experiments, it is not enough to cover all of IEDM needs. In particular, IEDM has to be able to describe actual experimental parameters such as the total fluence (number of particles received by unit of area) impinging on a piece of material during an IE. We rely on OM, the Units of Measure ontology, for the representation of experimental quantities and of their physical units [19]. OM contains entities from many physics-related domains, including concepts from particle physics. This ontology is thus necessary for the formalization of the physical quantities related to IEs and, as such, is the second most important ontology used for the development of IEDM.

### 2.3 FOAF

The third ontology from which we borrow to develop some concepts of IEDM is the Friend-of-a-Friend ontology (FOAF) [20]. This widely used ontology aims to describe networks of people, their activities, and their relations. For example, FOAF can be used to represent a social network on the web [21]. IEDM uses FOAF mostly to describe the characteristics of the various individuals involved in IEs.

## 3 IEDM Ontology Design Methodology

As mentioned above, in this work, we aim to maximize the reusability of upper-level ontologies that were described in Section 2. Therefore, we anchored the OWL-based definition of IEDM on the three foundational ontologies EXPO, OM and FOAF. These ontologies supplied the proper concepts, relations, and definitions for the representation of irradiation experiments, allowing for a more explicit taxonomy and axiomatization.

In IEDM, we reused EXPO for describing abstract and physical concepts linked to the various features of an irradiation experiment, from the definition of its requirements to the representation of its numerical experimental results. For example, we used the class `expo:AdminInfoExperiment` for representing the administrative information of an irradiation experiment. Yet, to ensure compatibility with the potential future enhancements of EXPO and other ontologies while ensuring also the compatibility with ontologies that are compliant with our three foundational ontologies (EXPO, OM and FOAF), one important IEDM design decision was to never copy, override or modify them. Instead, when more specific information was needed, we introduced

IEDM-specific variants of existing classes, using OWL namespaces to avoid ambiguities. Thus, as an example, we created `iedm:AdminInfoIrradiationExperiment` as a subclass of `expo:AdminInfoExperiment`, instead of directly updating the later.

To represent the experimental quantities specifically used when handling irradiation experiments, we relied on OM instead of EXPO. Entities such as `om:Energy`, `om:AbsorbedDose` (a dose is the amount of radiation energy delivered to matter per unit of mass) or `om:Activity` (an activity is a number of atom decays per unit of time) are fundamental concepts that appear in an irradiation experiment, and they exist in OM.

Another essential notion for our ontology is the concept of user. The term user has a broad definition and can be a group, an organization, or a person. Therefore, we employed the definition of `foaf:Agent`, which contains, as subclasses, all these three types of entities.

Of course, the domain of irradiation experiments is larger than those covered by the three foundational ontologies we used to integrate some of IEDM concepts. The second phase of the IEDM development thus required introducing concepts more specific to irradiation experiments per se. In this step, we followed a top-down approach: IE-specific concepts are added as subclasses of the upper ontologies' classes. For instance, `iedm:Element`, denoting the notion of atoms, is a subclass of `expo:Object`, and we added the notion of `iedm:RelativisticMomentum` or `iedm:Fluence`. We also added new relations, i.e., superclasses created via IE-specific OWL object properties. For example, the constraint `iedm:hasResult some iedm:CumulatedQuantity` is a superclass of the class `iedm:IrradiationExperiment`, which is, in fine, the only top-level new class introduced by IEDM.

During the integration of IEDM with the upper ontologies, we came across several issues. One of them was that a concept could appear in more than one of the three upper ontologies (e.g., `expo:Quantity` and `om:Quantity`). In that case, we decided to use the concept whose definition better fit our model. We also noticed cases where two concepts were in fact complementary to each other (e.g., `expo:Agent` and `foaf:Agent`), and hence we decided to create another IEDM class (`iedm:User`) that would be the subclass of both, taking advantage of OWL multiple inheritance for classes. Finally, to simplify the handling of IEDM-related instances, we only added one IEDM-specific data property, namely `iedm:hasValue`, which can be used at the OWL-instance level in a polymorphic fashion to assign data values to instances, for instance to specify actual numerical results (otherwise, as often, the OWL-generated label string is enough to represent the pertinent data).

## 4  IEDM Ontology Core Structure

In this section, we highlight, analyze, and explain the core entities and relations that characterize IEDM. Obviously, most of the entities that are described revolve around the Irradiation Experiment class (`iedm:IrradiationExperiment`). However, for easier comprehension, we start our description from the Irradiation Experiment Object class (`iedm:IrradiationExperimentObject`) and radiation field (`iedm:RadiationField`), which represent the key elements of an IE, and we build up our description to the `iedm:IrradiationExperiment` class (see Fig. 1).

### 4.1 Radiation Field

A radiation field (`iedm:RadiationField`) is necessary in order to perform an irradiation experiment. An `iedm:RadiationField` can be composed of particles (`iedm:Particle`) of the same type (`iedm:SingularField`) or more (`iedm:MixedField`).

### 4.2 Irradiation Experiment Object

An `iedm:IrradiationExperimentObject` object is a subclass of `expo:Object`; it represents an object of the irradiation facility infrastructure that is exposed to an `iedm:RadiationField` or an object that is under test. A Device Under Test (DUT) during an irradiation experiment is an instance of `iedm:DUT`. The purpose of performing an irradiation experiment is that the DUTs reach some target cumulated quantity (`iedm:CumulatedQuantity`), which is a subclass of `iedm:DosimetricQuantity`. Two of the most common cumulated quantities in irradiation experiments are the absorbed dose (`iedm:AbsorbedDose`) or fluence (`iedm:Fluence`), which are both included in our model.

### 4.3 Irradiation Experiment

An Irradiation Experiment (`iedm:IrradiationExperiment`) denotes a whole experiment where DUTs are exposed under a specific `iedm:RadiationField`. An instance of `iedm:IrradiationExperiment` has some explicit detailed specifications, which describe experimental methods, and includes a plan of actions that is defined by EXPO as `expo:ProcedureExecuteExperiment`. In IEDM, we extend this class with three subclasses. The first one is the `iedm:PassiveStandardIrradiation`, where an `iedm:DUT` is simply put into a `iedm:RadiationField`. The second class is `iedm:PassiveCustomIrradiation`, for which very specific `iedm:TechnicalRequirements` have to be specified and implemented by the experimenter. Finally, `iedm:ActiveIrradiation` represents the third category and includes experiments requiring an active data acquisition (DAQ) device and, typically, the usage of a readout system during the experiment execution. Each `iedm:IrradiationExperiment` instance belongs to one category of `expo:ProcedureExecuteExperiment`, which is defined by the OWL expression `iedm:hasIrradiationCategory exactly 1 expo:ProcedureExecuteExperiment`. As illustrated in Fig. 1, the `iedm:IrradiationExperiment` class has several relations represented as superclass expressions (e.g.,

```
expo:HasPart exactly 1 iedm:AdminInfoIrradiationExperi-
ment).
```

### 4.4 DUT Irradiation

The DUT Irradiation class (`iedm:DUTirradiation`) refers to a specific irradiation experiment related to a DUT. In contrast to `iedm:IrradiationExperiment`, this class depends on one and only one `iedm:DUT`, and it can be considered as part of an `iedm:IrradiationExperiment`. For this reason, we have defined the relation `iedm:IrradiationExperiment iedm:hasPart some iedm:DUTIrradiationExperiment`. The `iedm:DUTIrradiationExperiment` has two specific points in time, from `iedm:TimePosition`, that denote the start and the completion times of the radiation exposure.

### 4.5 Interaction Length and Occupancy

The parameters that define the degree of interaction of a particle with matter (interaction lengths) [22] are important quantities for an irradiation experiment. In an actual irradiation experiment as, for example, the one performed at CERN IRRAD (see Section 5), these values need to be kept as low as possible since a high degree of interaction (caused by too many DUTs placed in the `iedm:RadiationField` at the same time) can produce an excess of secondary particles that, in turn, can perturb the result of an `iedm:IrradiationExperiment`. These quantities, which are dependent upon the DUT materials and radiation fields and are computed by a dedicated physics-simulation program, are defined as `iedm:InteractionLength` and `iedm:InteractionLengthOccupancy`.

### 4.6 Element, Compound, and Layer

In our model, we include the concepts of elements (`iedm:Element`), corresponding to the basic entries of the Mendeleev periodic table, Compounds (`iedm:Compound`), which are mixtures of elements, and Layers (`iedm:Layer`), which are used to describe the 1D structure of a typical DUT along the `iedm:RadiationField` main propagation axis. These sets of information are necessary for properly computing quantities related to the `iedm:InteractionLength` and `iedm:InteractionLengthOccupancy`.

### 4.7 User

An `iedm:User` is defined as a subclass of both `expo:SentientAgent` and `foaf:Agent`. The `expo:SentientAgent` defines a user that has rights but may or may not have responsibilities and the ability to reason. If the latter is present, it can also have cognitive abilities. A `foaf:Agent` can be a person, a group, or an organization. By specifying that an `iedm:User` has both of them as superclasses, we can combine both concepts. In an irradiation facility, a user can have one or multiple roles during an irradiation experiment. These IEDM-specific roles are represented by subclasses of `expo:User`, which is an `expo:SubjectRole`, viewed in EXPO as a predicate, following SUMO's practice. In IEDM, we define the roles of irradiation facility coordinator (`iedm:IrradiationFacilityCoordinator`), irradiation facility manager (`iedm:IrradiationFacilityManager`) and irradiation facility user (`iedm:IrradiationFacilityUser`). The class `iedm:IrradiationFacilityUser` has two subclasses: the operator (`iedm:Operator`), who is a person that performs the irradiation experiment, and the responsible person (`iedm:ResponsiblePerson`), i.e., the person in charge of that specific irradiation experiment.

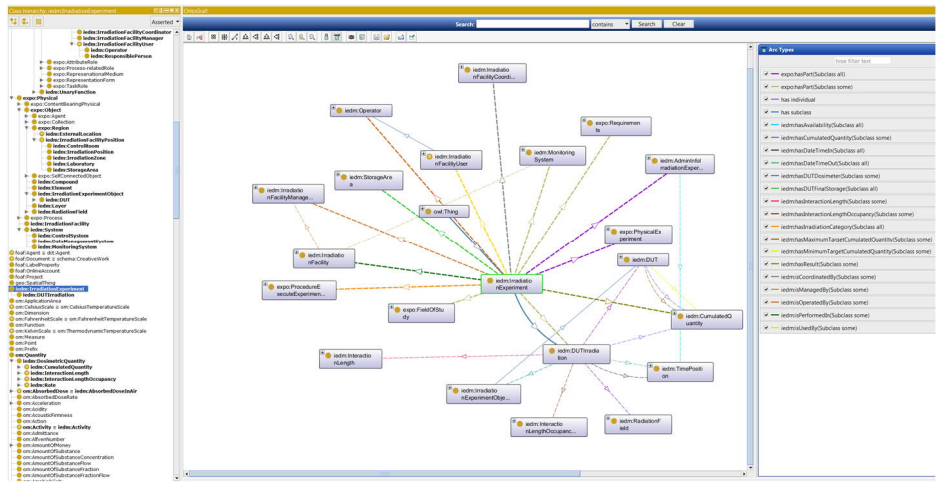

**Fig. 1.** Graph representation in Protégé of the IEDM ontology, with the focus on the `iedm:IrradiationExperiment` entity.

In the above subsections, we presented the main entities linked to an `iedm:IrradiationExperiment`. However, much more concepts exist in the IEDM ontology, in particular regarding the OWL object properties that relate these classes together. Therefore, we invite the interested reader to visit the IEDM Gitlab repository (https://gitlab.cern.ch/bgkotse/iedm.git) that contains all the resources, including copies of the version of EXPO, OM and FOAF used at the time of the definition of IEDM. All IEDM classes and relations are explained with proper annotations, definitions, and comments for easier comprehension.

## 5 IRRAD Use Case

To assess the validity, coherence, and completeness of the IEDM ontology, we applied it to the representation of a real irradiation experiment use case. CERN hosts the IRRAD proton irradiation facility, a research irradiation infrastructure that relies on the proton or ion beam delivered by the CERN Proton Synchrotron accelerator to perform IEs [11]. One of the IEs performed during the 2018 PS run was FCC-Radmon [9]. With the development of future accelerators such as the 100-Tev 80-kilometer-long Future Circular Collider [12]) being proposed by CERN, and its related experiments, targeting always higher performance, comes the need to monitor more intense radiation levels. This calls for the development of new generations of radiation monitors that can cope with these increased radiation levels. Thus, in the FCC-Radmon experiment, a new technology of particle fluence monitor was tested against radiation damage levels that could apply to a future accelerator such as the FCC [9].

Focusing mainly in the instances of the core classes (see Section 4), we aim to describe an IE with the concepts provided by IEDM. The name for this specific irradiation experiment instance is `iedm:FCC-Radmon`. For this experiment, we have created the `iedm:FCC-RadmonIrradiation` instance. The `iedm:FCC-RadmonIrradiation` represents an experiment of only one `iem:DUT`, which in this case is `iedm:PCB5-run2017`. For measuring, using a dedicated dosimeter, the actual `iedm:Fluence` during the `iedm:FCC-Radmon` experiment in the irradiation facility `iedm:CERN_IRRAD`, the `iedm:Operator1` has installed an `iedm:IrradiationExperimentObject` that, in this case, is the instance `iedm:Dosimeter004139`. The `iedm:FCC-RadmonIrradiation` instance has a specific start time (an `iedm:TimePosition`, which in our example is `iedm:_2018_03_30_12h_00`), time at which the DUT is put inside the PS proton radiation field (`iedm:Protons_24GeV`), and a specific end time, at which the radiation exposure is completed: `iedm:_2018_11_12_18h_00`. The `iedm:CumulatedQuantity` of interest, an `iedm:Fluence` in this experiment, is represented by the instance `iedm:_3e17_protons_per_square_cm` and has as `expo:MeasurementError iem:_7_per_cent`. (See Fig. 2). The complete use case can be found in the online resources (https://gitlab.cern.ch/bgkotse/iedm.git).

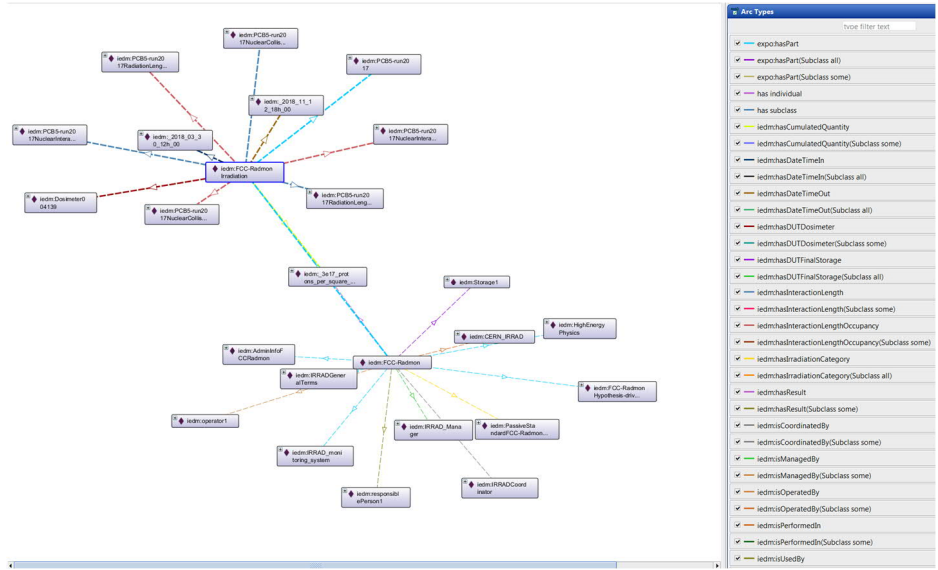

**Fig. 2.** Graph representation in Protégé of the FCC-Radmon IE instance.

## 6 Future Work

As detailed in Section 1, the motivations for the development of IEDM are many, thus leading to many possible future works, in addition to the maintenance and improvements that this first version of IEDM surely deserves. Beyond its use as a knowledge formalization foundation for the whole HEP IE community, a natural extension of the IEDM project would be to model, beyond irradiation experiments, a whole irradiation infrastructure, may be along the lines of the approach taken for designing ontologies for other large organizations such as utilities [23].

Yet, our current vision for the future is more oriented towards practical applications of IEDM. In fact, the conceptual model of IEDM has already been used as a guiding aide during the manual development of the IRRAD Data Manager (IDM) system [24], part of the AIDA-2020 project [13]. IDM has been successfully used for the handling and management of the data coming from the CERN IRRAD IEs during all of 2018 (see Fig. 3). Inspired by the similarities between IDM and the structure of IEDM definitions, we believe that the IEDM ontology itself can be the basis for the automatic generation of the user interface (UI) of data management systems for irradiation experiments, thus significantly shortening the development time of IT data management systems such as IDM. This application would also be, we believe, of benefit to other irradiation facilities, which often manage their experiments not in a standardized manner.

There have already been a few attempts to link ontologies and user interface design [25]. For our future work, we plan to provide a new approach in that regard, based on translating the IEDM classes into Python classes using the Owlready2 package [26]. This package allows loading OWL ontology classes, properties and instances as Python objects and manipulating them via Python code. By converting the IEDM entities into Python classes, we can use the ontology as the model for UI projects that follow a Model-View-Controller architecture (e.g., Django projects [27]). Our methodological emphasis, during the development of IEDM, on the clean separation of concerns between different ontologies and the focus on object properties always declared as subclasses, which naturally map to the object-oriented programming notion of attributes, should help streamline this translation process. Ultimately, a judicious design of IEDM-like ontologies in other fields should enable the automatic generation of the UI required by the corresponding data management systems.

**Fig. 3.** FCC-Radmon data as stored in IRRAD Data Manager.

## 7   Conclusion

We introduced IEDM, a new ontology for the data management of irradiation experiments. Such an ontology can be used for the formalization and standardization of knowledge related to irradiation experiments management, a concern for many of the hundreds of irradiation facilities in the world. IEDM allows for knowledge sharing and can be of great interest not only for the HEP scientific community, but also in other scientific fields such as medicine, space, etc. as well as for industrial applications.

More specifically, in this paper we described the methodology applied to build IEDM. Following a top-down approach, we based IEDM on three existing upper ontologies: EXPO, OM and FOAF. In a second step, these ontologies were extended and enriched with concepts relevant to irradiation experiments. The key classes and relations have also been explained, and an example of the IEDM encoding of an actual irradiation experiment, performed at the CERN IRRAD facility, is given, to illustrate the main notions behind IEDM and provide some evidence for the validity of its specification. Finally, we discussed the intriguing possibilities of using IEDM as a model for the automatic generation of user interfaces.

IEDM is open-source project and available at: https://gitlab.cern.ch/bgkotse/iedm.git.

## Acknowledgments

This project has received funding from the European Union's Horizon 2020 Research and Innovation program under Grant Agreement no. 654168 (AIDA-2020).